\documentclass{article}

\usepackage[final]{neurips_2020}

\usepackage[utf8]{inputenc} 
\usepackage[T1]{fontenc}    
\usepackage{url}            
\usepackage{booktabs}       
\usepackage{amsfonts}       
\usepackage{nicefrac}       
\usepackage{microtype}      

\usepackage{amsmath}
\usepackage{physics}
\usepackage{makecell}
\usepackage{graphicx}
\usepackage{algorithm,algorithmic}
\usepackage{multicol}
\usepackage{multirow}

\usepackage{wrapfig}
\usepackage{subfigure}
\usepackage{bm}
\usepackage{array, boldline}

\usepackage[svgnames, table]{xcolor}
\usepackage{qiangstyle}

\title{Firefly Neural Architecture Descent: a General Approach for Growing Neural Networks}

\author{%
  Lemeng Wu\thanks{Equal contribution.}\\
  Department of Computer Science\\
  University of Texas at Austin\\
  Austin, TX 78712 \\
  \texttt{lmwu@cs.utexas.edu} \\
  \And
  Bo Liu$^*$\\
  Department of Computer Science\\
  University of Texas at Austin\\
  Austin, TX 78712 \\
  \texttt{bliu@cs.utexas.edu} \\
  \And
  Peter Stone\\
  Department of Computer Science\\
  University of Texas at Austin\\
  Austin, TX 78712 \\
  \texttt{pstone@cs.utexas.edu} \\
  \And
  Qiang Liu\\
  Department of Computer Science\\
  University of Texas at Austin\\
  Austin, TX 78712 \\
  \texttt{lqiang@cs.utexas.edu} \\
}

\begin{document}
\maketitle

\begin{abstract}
We propose \emph{firefly neural architecture descent}, a general framework for progressively and dynamically growing neural networks to jointly optimize the networks' parameters and architectures.
Our method works in a steepest descent fashion, which iteratively finds the best network within a functional neighborhood of the original network
that includes a diverse set of candidate network structures.
By using Taylor approximation, the optimal network structure in the neighborhood can be found with a greedy selection procedure.
We show that firefly descent can flexibly grow networks both wider and deeper, and can be applied to learn accurate but resource-efficient neural architectures that avoid catastrophic forgetting in continual learning.
Empirically, firefly descent achieves promising results on both neural architecture search and continual learning. In particular, on a challenging continual image classification task, it learns networks that are smaller in size but have higher average accuracy than those learned by the state-of-the-art methods. The code is available at \url{https://github.com/klightz/Firefly}.
\end{abstract}

\section{Introduction}
Although biological brains are developed and shaped by complex progressive growing processes, most existing artificial deep neural networks are trained under fixed network structures (or architectures).
Efficient techniques that can progressively grow neural network structures can allow us to jointly optimize the network parameters and structures to achieve higher accuracy and computational efficiency, especially in dynamically changing environments.
%
For instance, it has been shown that accurate and energy-efficient neural network can be learned by progressively growing the network architecture starting from a relatively small network~\citep{splitting2019, wang2019energy}. Moreover, previous works also indicate that knowledge acquired from previous tasks can be transferred to new and more complex tasks by expanding the network trained on previous tasks to a functionally-equivalent larger network to initialize the new tasks~\citep{chen2015net2net, wei2016network}.

In addition, dynamically growing neural network has also been proposed as a promising approach for preventing the challenging \emph{catastrophic forgetting} problem in continual learning~\citep{rusu2016progressive, yoon2017lifelong, rosenfeld2018incremental, li2019learn}.

Unfortunately, searching for the optimal way to grow a network 
leads to a challenging combinatorial optimization problem. 
Most existing works use simple heuristics  \citep{chen2015net2net, wei2016network}, or  
 random search \citep{elsken2017simple, elsken2018efficient} to grow networks and may not fully unlock the power of network growing. 
An exception is  splitting steepest descent~\citep{splitting2019},
which considers growing networks by splitting the existing neurons into multiple copies, and derives a principled functional steepest-descent approach for 
determining which neurons to split and how to split them.
However, the method is restricted to neuron splitting, and can not incorporate more flexible ways for growing networks, including adding brand new neurons and introducing new layers.

In this work, we propose \emph{firefly neural architecture descent}, 
 a general and flexible framework for progressively growing neural networks.  
Our method is a local descent algorithm 
inspired by the typical gradient descent and  splitting steepest descent. 
It grows a network by  
 finding the best larger networks 
 in a \emph{functional neighborhood} of the original network whose size is controlled by a step size $\epsilon$, 
 which contains a rich set of networks that have various (more complex) structures, but are $\epsilon$-close to the original network in terms of the function that they represent.
 The key idea is that, when $\epsilon$ is small,  
 the combinatorial optimization on the
 functional neighborhood 
 can be simplified to a greedy selection, and therefore can be solved efficiently in practice.

The firefly neural architecture descent framework is highly flexible and practical
and allows us to derive general approaches for growing wider and deeper networks (Section~\ref{sec:grow-width}-\ref{sec:grow_layers}).  
It can be easily customized to address specific problems. For example, our method 
provides a powerful approach for dynamic network growing in continual learning (Section~\ref{sec:cl}), and 
can be applied to optimize cell structures in cell-based neural architecture search (NAS) such as DARTS \citep{liu2018darts} (Section~\ref{sec:expDarts}). Experiments show that Firefly  efficiently learns accurate and resource-efficient networks in various settings.
In particular, for continual learning, 
our method learns more accurate and smaller networks  
that can better prevent catastrophic forgetting, 
outperforming state-of-the-art methods such as Learn-to-Grow~\citep{li2019learn} and Compact-Pick-Grow~\citep{hung2019compacting}. 
\section{Firefly Neural Architecture Descent}
In this section, we start with introducing the general framework (Section~\ref{sec:general}) of firefly neural architecture descent. 
Then we discuss how the framework can be applied to grow a network both wider and deeper  (Section~\ref{sec:grow-width}-\ref{sec:grow_layers}). To illustrate the flexibility of the framework, 
we demonstrate how it can help tackle catastrophic forgetting in continual learning  (Section~\ref{sec:cl}). 
\subsection{The General Framework}
\label{sec:general} 
We start with the general problem of jointly optimizing neural network parameters and model structures.  
Let $\Omega$ be a space of neural networks with different parameters and structures (e.g., networks of various widths and depths). Our goal is to solve
\begin{align} \label{equ:Lf}
\argmin_{f}
\Big \{ L(f) ~~~s.t. ~~~ f\in \Omega,  ~~~ C(f) \leq \eta \Big\}, 
\end{align}
where $L(f)$ is the training loss function and $C(f)$ is a complexity measure of the network structure that reflects the computational or memory cost. This formulation poses a highly challenging optimization problem in a complex, hierarchically  structured space. 

We approach \eqref{equ:Lf} 
with a steepest descent type algorithm that generalizes typical parametric gradient descent and the splitting steepest descent of \citet{splitting2019}, 
with an iterative update of the form
\begin{align} \label{equ:masterupdate}
f_{t+1} = \argmin_{f} \Big \{ L(f) ~~~~s.t.~~~  f \in \neib(f_t, ~\epsilon),~~~~~~ 
C(f) \leq C(f_t)+\eta_{t} \Big\}, 
\end{align}
where we find the best network $f_{t+1}$ in 
neighborhood set $\neib(f_t, ~\epsilon)$ of the current network $f_t$ in $\Omega$, whose complexity cannot exceed that of $f_t$ by more than a threshold $\eta_t$. 
Here $\neib(f_t, ~\epsilon)$ denotes a neighborhood of  $f_t$ of ``radius'' $\epsilon$ such that $f(x) = f_t(x) +O(\epsilon)$ for $\forall f\in \neib(f_t, ~\epsilon)$. $\epsilon$ can be viewed as a small step size, which ensures that the network changes smoothly across iterations, and importantly, 
 allows us to use Taylor expansion to significantly simplify the optimization \eqref{equ:masterupdate} to yield practically efficient algorithms.  

The update rule in \eqref{equ:masterupdate} is highly flexible and reduces to different algorithms with different choices of $\eta_t$ and $\neib(f_t, \epsilon)$.
In particular, when $\epsilon$ is infinitesimal, by taking $\eta_t = 0$ and $\neib(f_t, \epsilon)$ the typical Euclidean ball on the parameters, \eqref{equ:masterupdate} reduces to standard gradient descent which updates the network parameters with architecture fixed.
However, by taking $\eta_t > 0$ and $\neib(f_t, \epsilon)$ a rich set of neural networks with different, larger network structures than $f_t$, we obtain novel \emph{architecture descent} rules that allow us to incrementally grow networks. 
   
   In practice, we alternate between parametric descent and architecture descent according to a user-defined schedule (see Algorithm~\ref{alg:main}). Because architecture descent increases the network size, it is called less frequently (e.g., only when a parametric local optimum is reached). From the optimization perspective, 
   performing architecture descent allows us to lift the optimization into a higher dimensional space with more parameters, and hence escape local optima that cannot be escaped in the lower dimensional space (of the smaller models). 
   
   In the sequel, we instantiate the neighborhood $\neib(f_t, ~\epsilon)$ for growing wider and deeper networks, and for continual learning, and discuss how to solve the optimization in \eqref{equ:masterupdate} efficiently in practice. 
 
\begin{algorithm*}[t] 
    \label{alg:main}
    \caption{Firefly Neural Architecture Descent}  
    \begin{algorithmic} 
        \STATE \textbf{Input}: 
        Loss function $L(f)$; 
        initial small network $f_0$; 
        search neighborhood $\neib(f, \epsilon)$; 
        maximum increase of size $\{\eta_t\}$. 
        \STATE \textbf{Repeat:} At the $t$-th growing phase:
            \STATE \textbf{1.} Optimize the parameter of $f_t$ with fixed structure using a typical optimizer for several epochs.
            \STATE \textbf{2.} Minimize $L(f)$ in $f \in \neib(f, \epsilon)$ without the complexity constraint (see e.g., \eqref{equ:wideroptwithno}) to get a large ``over-grown'' network $\tilde f_{t+1}$ by performing gradient descent.
            \STATE \textbf{3.}  Select the top $\eta_t$ neurons in $\tilde f_{t+1}$ with the highest importance measures to get $f_{t+1}$ (see \eqref{equ:epsilonoptwide}).  
    \end{algorithmic}
    \label{alg:firefly}  
\end{algorithm*}

\subsection{Growing Network Width}
\label{sec:grow-width} 
\begin{figure}[t]
    \centering
    \includegraphics[width=\textwidth]{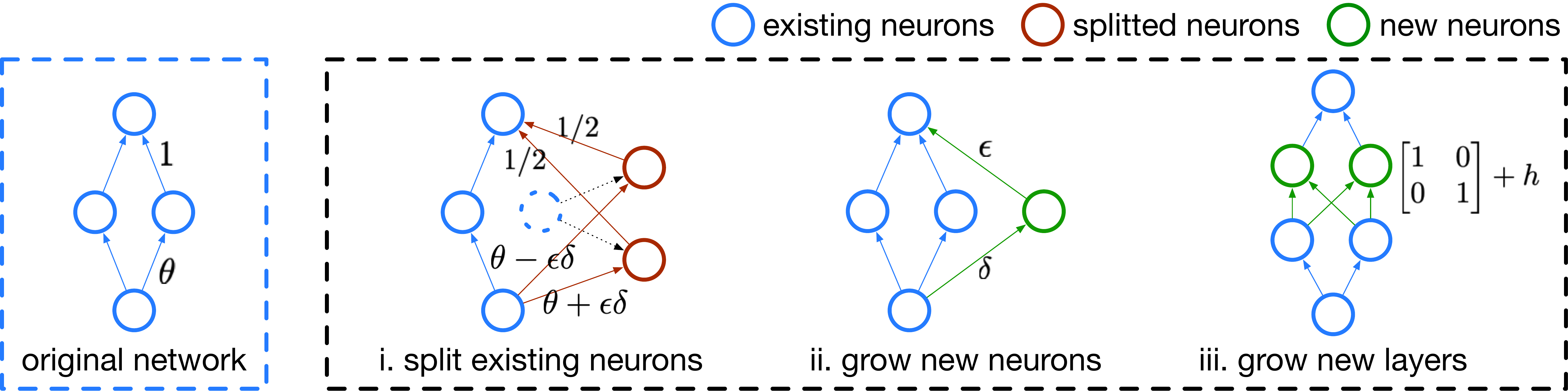}
    \caption{An illustration of three different growing methods within firefly neural architecture descent. Both $\delta$ and $h$ are trainable perturbations.}
    \label{fig:firefly_method}
\end{figure}

We discuss how to define $\neib(f_t, \epsilon)$ to progressively build increasingly wider networks, and then introduce how to efficiently solve the optimization in practice.   
We illustrate the idea with two-layer networks, but extension to multiple layers works straightforwardly. 
Assume $f_t$ is a two-layer neural network (with one hidden layer) of the form  $f_t(x) = \sum_{i=1}^{m} \sigma(x, \theta_i)$, 
where $\sigma(x, \theta_i)$ denotes its $i$-th neuron with parameter $\theta_i$ and $m$ is the number of neurons (a.k.a. width).
There are two ways to introduce new neurons 
to build a wider network, 
including splitting existing neurons in $f_t$ and introducing brand new neurons; see Figure~\ref{fig:firefly_method}. 
\paragraph{Splitting Existing Neurons} Following \citet{splitting2019},
an essential approach to growing neural networks is to 
split the neurons into a linear combination of multiple similar neurons. 
Formally, 
splitting a neuron $\theta_i$\footnote{A neuron is determined by both $\sigma$ and $\theta$. But since $\sigma$ is fixed under our discussion, we abuse the notation and use $\theta$ to represent a neuron.} into a set of neurons $\{\theta_{i\ell}\}$ with weights $\{w_{i\ell}\}$ 
amounts to replacing $\sigma(x, \theta_i)$ in $f_t$ with $\sum_{\ell } w_{i\ell} \sigma(x, \theta_{i\ell})$. 
We shall require that 
$\sum_{\ell} w_{i\ell} =1$ and $\norm{\theta_{i\ell} - \theta}_2 \leq \epsilon,$ $\forall \ell$ so that the new network is $\epsilon$-close to the original network. 
As argued in \citet{splitting2019}, when $f_t$ reaches a parametric local optimum and $w_{i\ell} \geq 0$, 
it is sufficient to consider a simple binary splitting scheme, which splits a neuron $\theta_i$ into two equally weighted copies along opposite update directions, that is, 
$\sigma(x, \theta_i) 
\Rightarrow 
 \frac{1}{2} \big ( \sigma(x, \theta_i + \epsilon \delta_{i}) 
 + \sigma(x, \theta_i -  \epsilon \delta_i)     
\big )
$, where $\delta_i$ denotes the update direction.

\paragraph{Growing New Neurons}  Splitting the existing neurons yields a ``local'' change 
because the parameters of the new neurons are close to that of the original neurons. 
A way to introduce ``non-local'' updates is to add brand new neurons with arbitrary parameters far away from the existing neurons. 
This is achieved by replacing $f_t$ with 
$f_t(x) + \epsilon\sigma(x, \delta)$, 
where $\delta$ now denotes a trainable parameter of the new neuron and the neuron is multiplied by $\epsilon$ to ensure the new network is close to $f_t$ in function. 

Overall, 
to grow $f_t(x) = \sum_i \sigma(x; \theta_i)$ wider, 
the neighborhood set $\neib(f_t, \epsilon)$ can include functions of the form
$$
f_{\vv\varepsilon,\vv\delta}(x) = \sum_{i=1}^m \frac{1}{2}\Big (\sigma(x, \theta_i +\varepsilon_i\delta_i)  
+ \sigma(x, \theta_i - \varepsilon_i \delta_i) \Big ) + \sum_{i=m+1}^{m+m'} \varepsilon_i \sigma(x, \delta_i),
$$
where we can potentially split all the neurons in $f_t$ and add upto $m'$ new non-local neurons ($m'$ is a hyperparameter). 
Whether each new neuron will eventually be added is  controlled by an individual step-size $\varepsilon_i$ that satisfies $|\varepsilon_i| \leq \epsilon$. If $\varepsilon_i = 0$, it means the corresponding new neuron is not introduced. 
Therefore, the number of new neurons introduced in  $f_{\vv\varepsilon, \vv\delta}$  equals the $\ell_0$ norm 
$\norm{\vv\varepsilon}_0 := \sum_{i=1}^{m+m'} \ind(\varepsilon_i = 0)$. 
Here  $\vv\varepsilon = [\varepsilon_i]_{i=1}^{m+m'}$ and $\vv\delta = [\delta_i]_{i=1}^{m+m'}$. 

Under this setting, the optimization in \eqref{equ:masterupdate} can be framed as 
\begin{align} \label{equ:wideropt}
\min_{\vv\varepsilon, \vv\delta}\Big\{  L(f_{\vv\varepsilon, \vv\delta}) ~~~s.t.~~~ \norm{\vv\varepsilon}_0 \leq \eta_{t},~~~\norm{\vv\varepsilon}_\infty \leq \epsilon,~~~
\norm{\vv\delta}_{2,\infty} \leq 1 \Big\},
\end{align}
where $\norm{\vv\delta}_{2,\infty}=\max_i \norm{\delta_i}_2$, which is constructed to prevent $\norm{\delta_i}_2$ from becoming arbitrarily large. 
\paragraph{Optimization} It remains to solve the optimization in \eqref{equ:wideropt}, 
which is challenging due to the $\ell_0$ constraint on $\vv\varepsilon$. 
However, when the step size $\epsilon$ is small,  
we can solve it approximately with a simple two-step method: we first optimize $\vv \delta$ and $\vv\varepsilon$ while dropping the $\ell_0$ constraint, and then re-optimize $\vv\varepsilon$ with Taylor approximation on the loss, which amounts to simply picking the new neurons with the largest contribution to the decrease of loss, measured by the  gradient magnitude. 

\emph{Step One.} Optimizing $\vv \delta$ and $\vv\varepsilon$ without the sparsity constraint $\norm{\vv\varepsilon}_0
\leq \eta_t$, that is,  
\begin{align} \label{equ:wideroptwithno}
[\tilde{\vv\varepsilon}, \tilde{\vv\delta}]=\argmin_{\vv\varepsilon, \vv\delta} \Big \{ L(f_{\vv\varepsilon, \vv\delta})
~~~s.t.~~~ 
\norm{\vv\varepsilon}_\infty \leq \epsilon,~~~
\norm{\vv\delta}_{2,\infty} \leq 1 \Big\}.
\end{align} 
In practice, we solve the optimization with gradient descent by 
turning the constraint into a penalty. 
Because $\epsilon$ is small, we only need to perform a small number of gradient descent steps.  

\emph{Step Two.} Re-optimizing $\vv\varepsilon$ with Taylor approximation on the loss. To do so, note that when $\epsilon$ is small, we have by Taylor expansion: 
$$
L(f_{\vv\varepsilon, \tilde{\vv\delta}})  = L(f)
 ~+~ \sum_{i=1}^{m+m'}\varepsilon_i s_i ~+~ O(\epsilon^2),~~~~~~~ 
s_i = \frac{1}{\tilde\varepsilon_i}\int_{0}^{\tilde \varepsilon_i}\nabla_{\zeta_i} L(f_{[\tilde{\vv\varepsilon}_{\neg i}, \zeta_i], \tilde{\vv\delta}}) d \zeta_i,
$$
where $[\tilde{\vv\varepsilon}_{\neg i}, \zeta_i]$ denotes 
replacing the $i$-th element of $\tilde{\vv\varepsilon}$ with $\zeta_i$, and $s_i$ is an integrated gradient that measures the contribution of turning on the $i$-th new neuron. In practice, we approximate the integration in $s_i$ by discrete sampling:  
 $s_i \approx \frac{1}{n} \sum_{z=1}^{n} \nabla_{c_z} L(f_{[\tilde{\vv\varepsilon}_{\neg i}, c_z], \tilde{\vv\delta}})$ with $c_z =({2z-1})/{2n} \tilde{\varepsilon}_i$ and $n$ a small integer (e.g., $3$). 
Therefore, optimizing $\vv\varepsilon $ with fixed $\vv\delta = \tilde{\vv\delta}$ can be approximated by 
\begin{align} \label{equ:epsilonoptwide}
\hat{\vv\varepsilon} =\argmin_{ \vv\varepsilon } \Big\{ \sum_{i=1}^{m+m'} \varepsilon_i s_i  ~~~~s.t.~~~ \norm{\vv\varepsilon}_0 \leq \eta_{t},~~~ 
\norm{\vv\varepsilon}_\infty \leq \epsilon
\Big\}. 
\end{align}
It is easy to see that finding the optimal solution reduces to selecting the neurons with the largest gradient magnitude $|s_i|$. Precisely, we have $
\hat{\varepsilon}_i = -\epsilon~\ind(|s_i| \geq|s_{(\eta_t)}|)~ \sign(s_i)$, where $|s_{(1)}| \leq |s_{(2)}| \leq \cdots$ is the increasing ordering of $\{|s_i|\}$. 
Finally, we take $f_{t+1} = f_{\hat{\vv\varepsilon},\tilde{\vv\delta}}.$ 

It is possible to further re-optimize $\vv\delta$ with fixed $\vv \varepsilon $ and repeat the alternating optimization iteratively. However, performing the two steps above is computationally efficient and already solves the problem reasonably well as we observe in practice. 

\paragraph{Remark} When we include only neural splitting in $\neib(f_t, \epsilon)$, our method is equivalent to  
splitting steepest descent~\citep{splitting2019}, but with a simpler and more direct gradient-based optimization rather than solving the eigen-problem in~\citet{splitting2019, wang2019energy}. 

\subsection{Growing New Layers}
\label{sec:grow_layers}
We now introduce how to grow new layers under our framework. 
The idea is to include in $\neib(f_t, \epsilon)$ 
deeper networks with extra trainable residual layers 
and to select the layers (and their neurons) that contribute the most to decreasing the loss using the similar two-step method described in Section~\ref{sec:grow-width}.

Assume $f_t$ is a $d$-layer deep neural network of form $f_t = g_d \circ \cdots \circ g_{1}$, where $\circ$ denotes function composition. 
In order to grow new layers, 
we include in $\neib(f_t, \epsilon)$ functions of the form 
\begin{align*} 
f_{\vv\varepsilon, \vv\delta} = g_d \circ (I +  h_{d-1}) \cdots  (I + h_2)\circ  g_2\circ (I+  h_1) \circ g_{1}, &&\text{with}&& h_{\ell}(\cdot)= \sum_{i=1}^{m'} \varepsilon_{\ell i} \sigma(\cdot, \delta_{\ell i}),
\end{align*} 
in which we insert new residual layers of form $I + h_{\ell}$; here $I$ is the identity map,  and $h_{\ell}$ is a layer that can consist of upto $m'$ newly introduced neurons. 
Each neuron in 
$h_\ell$ is associated with a trainable parameter $\delta_{\ell i}$ and multiplied by $\varepsilon_{\ell i} \in [-\epsilon,\epsilon]$. 
As before,
the $(\ell i)$-th neuron is turned off if  $\varepsilon_{\ell i} = 0$,  and 
the whole layer $h_\ell$ is turned off if $\varepsilon_{\ell i} =0$ for all $i\in [1,m']$.  
Therefore, 
the number of new neurons introduced in $f_{\vv\varepsilon, \vv\delta}$ equals $\norm{\vv\varepsilon}_{0} := \sum_{i\ell} \ind(\epsilon_{i\ell}\neq 0)$, and the number of new layers added equals 
$\norm{\vv\varepsilon}_{\infty,0}: =\sum_{\ell} \ind(\max_i |\varepsilon_{\ell i}| \neq 0)$. 
Because adding new neurons and new layers 
have different costs, they can be controlled by two separate budget constraints (denoted by $\eta_{\eta_{t,0}}$ and $\eta_{t,1}$, respectively). 
Then the optimization of the new network can be framed as 
$$
\min_{\vv\varepsilon, \vv\delta} \Big\{ L(f_{\vv\varepsilon, \vv\delta})  ~~~s.t.~~~ \norm{\vv\varepsilon}_{0} \leq \eta_{t,0}, ~~  \norm{\vv\varepsilon}_{\infty,0} \leq \eta_{t,1},~~~
\norm{\vv\varepsilon}_\infty\leq \epsilon, ~~~
\norm{\vv\delta}_{2,\infty} \leq 1\Big\}, 
$$
where $\norm{\vv\delta}_{2,\infty} = \max_{\ell,i}\norm{\delta_{\ell i}}_2$. 
This optimization can be solved with a similar two-step method to the one for growing width, as described in Section \ref{sec:grow-width}: 
we first find the optimal $[\tilde{\vv\epsilon},\tilde{\vv\delta}]$ without the 
complexity constraints (including $\norm{\vv\varepsilon}_{0} \leq \eta_{t,0},$  $\norm{\vv\varepsilon}_{0,\infty} \leq \eta_{t,1}$), and then re-optimize $\vv\varepsilon$ with a Taylor approximation of the objective: 
\begin{align*} 
\min_{\vv\varepsilon} \bigg\{\sum_{\ell i} \epsilon_{\ell i} s_{\ell i}~~~s.t.~~~ \norm{\vv\varepsilon}_{0} \leq \eta_{t,0}, ~~  \norm{\vv\varepsilon}_{\infty,0} \leq \eta_{t,1}\bigg\}, ~
\text{where}~ s_{\ell i} = \frac{1}{\tilde\varepsilon_{\ell  i}}\int_{0}^{\tilde \varepsilon_{\ell i}} \!\!\!\!\!\nabla_{\zeta_{\ell i}} L(f_{[\tilde{\vv\varepsilon}_{\neg \ell i}, \zeta_{\ell i}], \tilde{\vv\delta}}) d \zeta_{\ell i}. 
\end{align*}
The solution can be obtained by sorting $|s_{ti}|$ in descending  order and selecting the top-ranked neurons until the complexity constraint is violated. 

\paragraph{Remark} In practice, we can apply all methods above to simultaneously grow the network wider and deeper. Firefly descent can also be extended to various other growing settings without case-by-case mathematical derivation. Moreover, the space complexity to store all the intermediate variables is $\mathcal{O}(N+m')$, where $N$ is the size of the sub-network we consider expanding and $m'$ is the number of new neuron candidates.\footnote{Because all we need to store is the gradient, which is of the same size as the original parameters.}


\subsection{Growing Networks in Continual Learning} 
\label{sec:cl}
Continual learning (CL) studies the problem of learning a sequence of different tasks (datasets) that arrive in a temporal order,
so that whenever the agent is presented with a new task, it no longer has access to the previous tasks.
As a result, one major difficulty of CL is to avoid \emph{catastrophic forgetting}, in that learning the new tasks severely interferes with the knowledge learned previously and causes the agent to ``forget'' how to do previous tasks. One branch of approaches in CL consider dynamically growing networks to avoid \emph{catastrophic forgetting}~\citep{rusu2016progressive,li2017learning,yoon2017lifelong,li2019learn,hung2019compacting}. However, most existing growing-based CL methods 
use hand-crafted rules to expand the networks (e.g. uniformly expanding each layer) and do not explicitly seek for the best growing approach under a principled optimization framework.
We address this challenge with the Firefly architecture descent framework.

Let $\D_t$ be the dataset appearing at time $t$ and $f_t$ be the network trained for $\D_t$. 
At each step $t$, we maintain a master network $f_{1:t}$ consisting of the union of all the previous networks $\{f_{s}\}_{s=1}^t$, such that each $f_s$ can be retrieved by applying a proper binary mask. 
When a new task $\D_{t+1}$ arrives, 
we construct $f_{t+1}$ by leveraging the existing neurons in $f_{1:t}$ as much as possible, 
while adding a controlled number of new neurons 
to capture the new information in $\D_{t+1}.$

\begin{figure}[t]
    \centering
    \includegraphics[width=\textwidth]{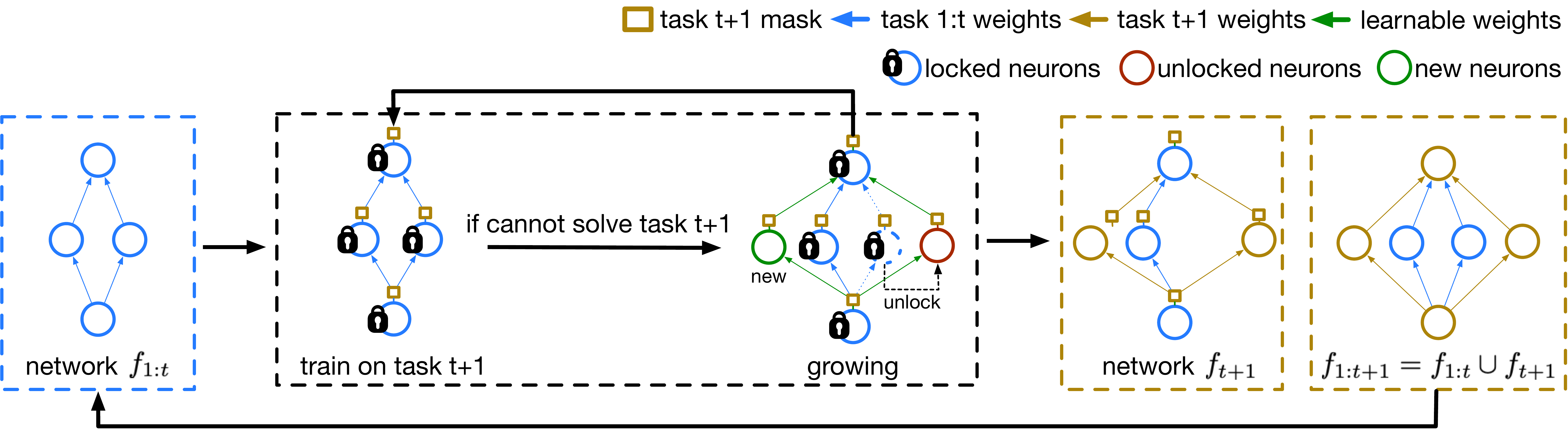}
    \caption{Illustration of how Firefly grows networks in 
     continual learning.}
    \label{fig:cl_grow}
    \vspace{-10pt}
\end{figure}

Specifically, we design 
$f_{t+1}$ to include three types of neurons 
(see Figure~\ref{fig:cl_grow}): \textbf{1)} \emph{Old neurons from $f_{1:t}$}, whose parameters are \emph{locked} during the training of $f_{t+1}$ on the new task $\D_{t+1}$. This does not introduce extra memory cost.
\textbf{2)} \emph{Old neurons from $f_t$}, whose parameters are \emph{unlocked and updated} during the training of $f_{t+1}$ on $\D_{t+1}$. This introduces new neurons and hence increases the memory size. It is similar to network splitting in Section~\ref{sec:grow-width} in that the new neurons are evolved from an old neuron, but only one copy is generated and the original neuron is not discarded. 
\textbf{3)} \emph{New neurons} introduced in the same way as in Section~\ref{sec:grow-width},\footnote{It is also possible to introduce new layers for continual learning, which we leave as an interesting direction for future work.} which also increases the memory cost.
Overall, assuming $f_{1:t}(x)=\sum_{i=1}^m \sigma(x; \theta_i)$, possible candidates of $f_{t+1}$ indexed by $\vv\varepsilon,\vv\delta$ are of the form:
$$
f_{\vv\varepsilon, \vv\delta} (x)= 
\sum_{i=1}^m \sigma(x; \theta_i + \varepsilon_i \delta_i) + 
\sum_{i=m+1}^{m+m'} \varepsilon_i \sigma(x; \delta_i), 
$$
where $\varepsilon_i \in [-\epsilon, \epsilon]$ again 
controls if the corresponding neuron is locked or unlocked (for $i\in [m]$), or if the new neuron should be introduced (for $i>m$). The new neurons introduced into the memory are $\norm{\vv\varepsilon}_0 = \sum_{i=1}^{m+m'} \ind(\varepsilon \neq 0)$.
The optimization of $f_{t+1}$ can be framed as 
$$
f_{t+1}=\argmin_{\vv\varepsilon, \vv\delta}\Big\{ L(f_{\vv\varepsilon, \vv\delta}; ~ \D_{t+1}) ~~~~s.t.~~~~ \norm{\vv\varepsilon}_0\leq \eta_t, ~~ \norm{\vv\varepsilon}_\infty \leq \epsilon, ~~~ \norm{\vv\delta}_{2,\infty}\leq 1\Big\},
$$
where $L(f; \mathcal D_{t+1})$ denotes the training loss on dataset $\D_{t+1}$. 
The same two-step method  in Section \ref{sec:grow-width} can be applied to solve the optimization. After $f_{t+1}$ is constructed, the new master network $f_{1:t+1}$ is constructed by merging $f_{1:t}$ and $f_{t+1}$ and the binary masks of the previous tasks are updated accordingly. 
See Appendix \ref{app:firefly} for the detailed algorithm.

\section{Empirical Results}\label{sec:experiments}
We conduct four sets of experiments to verify the effectiveness of firefly neural architecture descent.  
In particular, we first demonstrate the importance of introducing additional growing operations beyond neuron splitting \citep{splitting2019} 
and then apply the firefly descent to both neural architecture search and continual learning problems. In both applications, firefly descent finds competitive but more compact networks in a relatively shorter time compared to state-of-the-art approaches.

\begin{figure}[t]
    \centering
    \includegraphics[width=\textwidth]{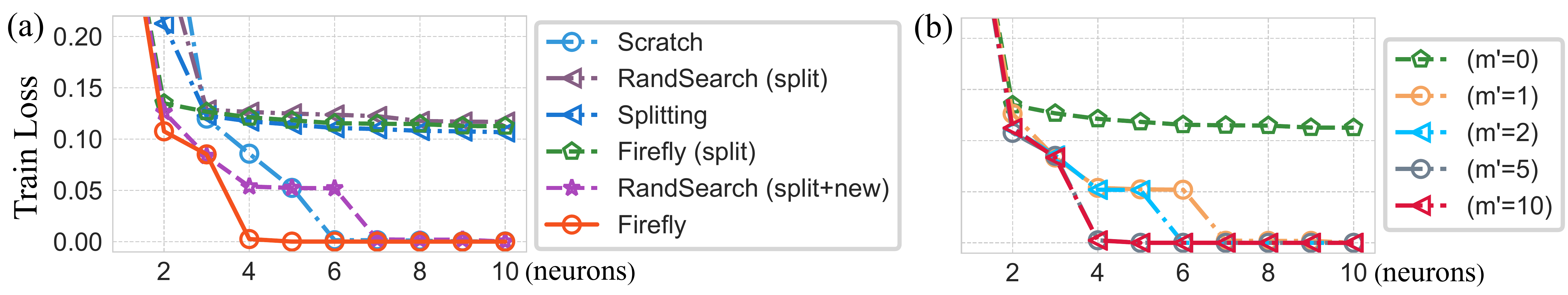}
    \caption{(a) Average training loss of different growing methods versus the number of grown neurons. (b) Firefly descent with different numbers of new neuron candidates.}
    \label{fig:toy}
    \vspace{-10pt}
\end{figure}

\paragraph{Toy RBF Network} \label{sec:toy} We start with growing a toy single-layer network to demonstrate the importance of introducing brand new neurons over pure neuron splitting. In addition, we show the local greedy selection in firefly descent is efficient by comparing it against random search.
Specifically, we adopt a simple two-layer radial-basis function (RBF) network with one-dimensional input and compare various methods that grow the network gradually from 1 to 10 neurons. 
The training data consists of 1000 data points from a randomly generated RBF network. 
We consider the following methods: 
{\tt Firefly}:  firefly descent for growing wider by splitting neuron and adding upto $m'=5$ brand new neurons; 
{\tt Firefly (split)}:  
 firefly descent for growing wider with only neuron splitting (e.g., $m'=0$);  
 {\tt Splitting}: the steepest splitting descent of \citet{splitting2019}; 
{\tt RandSearch (split)}: randomly selecting one neuron and splitting in a random direction, repeated $k$ times to pick the best as the actual split; we take $k=3$ to match the time cost with our method;
{\tt 
RandSearch (split+new)}: the same as  {\tt RandSearch (split)} but with 5 randomly initialized brand new neurons in the candidate during the random selecting;
{\tt Scratch}: training networks with fixed structures starting from scratch. 
We repeat each experiment 20 times with different ground-truth RBF networks and report the mean training loss in Figure~\ref{fig:toy}(a).

As shown in Figure~\ref{fig:toy}~(a), the methods with pure neuron splitting (without adding brand new neurons) can easily get stuck at a relatively large training loss and splitting further does not help escape the local minimum.  
In comparison, all methods that introduce additional brand new neurons can optimize the training loss to zero. Moreover, {\tt Firefly} grows neural network the better than random search under the same candidate set of growing operations.

We also conduct a parameter sensitivity analysis on $m'$ in Figure~\ref{fig:toy}(b), which shows the result of {\tt Firefly} as we change the number $m'$ of the brand new neurons. We can see that the performance improves significantly by even just adding one brand new neuron in this case, and the improvement saturates when $m'$ is sufficiently large ($m'=5$ in this case).

\paragraph{Growing Wider and Deeper Networks}\label{sec:expwider} We test the effectiveness of firefly descent for both growing network width and depth.
We use VGG-19~\citep{simonyan2014very} as the backbone network structure and compare our method with splitting steepest descent \citep{splitting2019}, 
Net2Net \citep{chen2015net2net} which grows networks uniformly by randomly selecting the existing neurons in each layer,
and neural architecture search by hill-climbing (NASH) \citep{elsken2017simple}, which is a random sampling search method using network morphism on CIFAR-10. 
For Net2Net, the network is initialized as a thinner version of VGG-19,  
whose layers are
$0.125\times$ the original sizes. 
For splitting steepest descent, NASH, and our method,  
we initialize the VGG-19 with 16 channels in each layer.  
For firefly descent,  we grow a network by both splitting  existing neurons and adding brand new neurons for widening the  network;  
we add $m'=50$ brand new neurons  
and set the budget to grow the size by $30\%$ at each step of  our method. 
See Appendix \ref{app:grow} for more information on the setting. 

\begin{figure}[h]
    \centering
    \includegraphics[width=\textwidth]{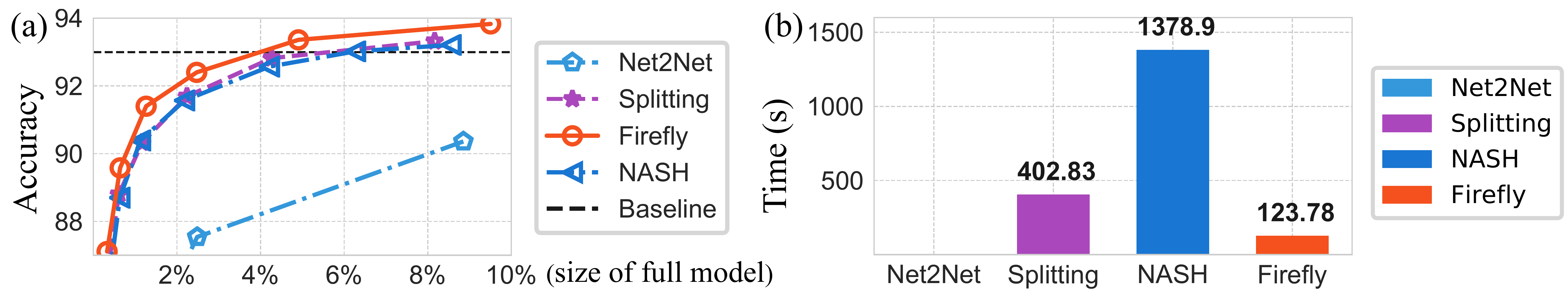}
    \caption{(a) Results of growing increasingly wider networks on CIFAR-10; VGG-19 is used as the backbone. (b) Computation time spent on growing for different methods.
    }
    \label{fig:cifar}
\end{figure}
%

    

Figure~\ref{fig:cifar} (a) shows the test accuracy, where the x-axis is the percentage of the grown model's size over the standard VGG-19.
We can see that 
the proposed method clearly outperforms the \emph{splittting steepest descent} and  Net2Net.
In particular, we 
achieve comparable test accuracy as the full model with only $4\%$ of the full model's size. Figure~\ref{fig:cifar}(b) shows the average time cost of each growing method for one step, we can see that {\tt Firefly} performs much faster than splitting the steepest descent and NASH. 
We also applied our method 
to gradually grow new layers in neural networks, 
we compare our method with NASH \citep{elsken2017simple} and AutoGrow \citep{wen2019autogrow}.
Due to the page limit, we defer the detailed results to Appendix \ref{app:grow}.

\paragraph{Cell-Based Neural Architecture Search}
\label{sec:expDarts} Next, we apply our method as a new way for improving cell-based Neural Architecture Search (NAS) \citep[e.g.][]{  zoph2018learning, liu2018progressive, real2019regularized}. 
The idea of cell-based NAS is to learn optimal neural network modules (called cells), from a predefined search space, 
such that they serve as good building blocks 
to composite complex neural networks. Previous works mainly focus on using reinforcement learning or gradient based methods to learn a sparse cell structure from a predefined parametric template. Our method instead gradually grows a small parametric template during training and obtains the final network structure according to the growing pattern.

Following the setting in DARTS \citep{liu2018darts}, 
we build up the cells as computational graphs whose structure is the directed DAG with 7 nodes.
%
The edges between the nodes
are linear combinations of different computational operations (SepConv and DilConv of different sizes) and the identity map. 
To grow the cells, 
we apply firefly descent to grow 
the number of channels in each operation by both splitting existing neurons and adding brand new neurons. 
%
During search, we compose a network by 
stacking 5 cells sequentially to evaluate the quality of the cell structures. We train 100 epochs in total for searching, and grow the cells every 10 epochs. After training, the operation with the largest number of channels on edge is selected into the final cell structure. In addition, if the operations on the same edge all only grow a small amount of channels compared with the initial setting, we select the Identity operation instead. The network that we use in the final evaluation is a larger network consisting of 20 sequentially stacked cells.  
More details of the experimental setup can be found in Appendix \ref{app:nas}.

Table \ref{tab:nas} reports the results 
comparing {\tt Firefly} with several NAS baselines. 
Our method achieves a similar or better performance comparing with those RL-based and gradient-based methods like ENAS or DARTS, but with higher computational efficiency in terms of the total search time.
\begin{table}[h]
\parbox{1.0\linewidth}{
\renewcommand\arraystretch{1.0}
    \centering
    \scalebox{0.95}{
    \begin{tabular}{c|c cc}
    \hlineB{2.5}
        Method & Search Time (GPU Days) & Param (M) & Error  \\
         \hline
         NASNet-A \citep{zoph2018learning}& 2000 & 3.1 & $2.83$
         \\
         ENAS \citep{pham2018efficient} & 4 & 4.2 & $2.91$
         \\
         Random Search & 4 & 3.2 & $3.29 \pm 0.15$
         \\
      
          DARTS (first order) \citep{liu2018darts} & 1.5 & 3.3 & $3.00 \pm 0.14 $ \\

          DARTS (second order) \citep{liu2018darts}  & 4 & 3.3 & $ 2.76 \pm 0.09$ \\

          {\tt Firefly} & 1.5 & 3.3 & $ 2.78 \pm 0.05 $  \\
    \hlineB{2.5}
    \end{tabular}
    }
    \vspace{5pt}
    \caption{Performance compared with several NAS baseline}
    \label{tab:nas}
   }
   \vspace{-15pt}
\end{table}

\begin{figure}[t!]
    \centering
    \begin{tabular}{c}
         \includegraphics[width=0.95\textwidth]{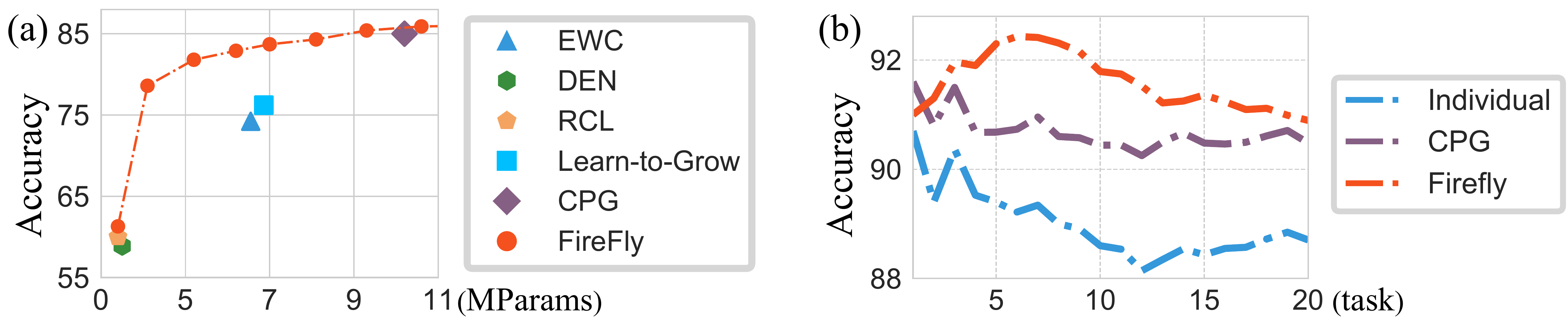}
    \end{tabular}
    \caption{(a) Average accuracy on 10-way split of CIFAR-100 under different model size. We compare against Elastic Weight Consolidation (EWC)~\citep{kirkpatrick2017overcoming}, Dynamic Expandable Network (DEN)~\citep{yoon2017lifelong}, Reinforced Continual Learning (RCL)~\citep{xu2018reinforced} and Compact-Pick-Grow (CPG)~\citep{hung2019compacting}. (b) Average accuracy on 20-way split of CIFAR-100 dataset over 3 runs. Individual means train each task from scratch using the Full VGG-16.}
    \label{fig:cl}
    \vspace{-10pt}
\end{figure}

\textbf{Continual Learning}
\label{sec:expCL}~~
Finally, we apply our method to grow networks  for  continual learning (CL), and compare with two state-of-the-art methods,  Compact-Pick-Grow (CPG) \citep{hung2019compacting} and  Learn-to-grow~\citep{li2019learn},  
both of which also progressively grow neural networks for learning  new tasks.
For our method, we grow the networks starting from a thin variant of the original VGG-16 without fully connected layers. 

Following the setting in Learn-to-Grow, we construct 10 tasks by randomly partitioning CIFAR-100 into 10 subsets. 
Figure~\ref{fig:cl}(a) shows the average accuracy and size of models at the end of the 10 tasks learned by firefly descent, Learn-to-Grow, CPG and other CL baselines.
We can see that firefly descent learns smaller networks with higher accuracy. 
To further compare with CPG, we follow the setting of their original paper \citep{hung2019compacting} and randomly partition CIFAR-100 to 20 subsets of 5 classes to construct 20 tasks.
Table~\ref{tab:cl20} shows the average accuracy and size learned at the end of 20 tasks. Extra growing epochs refers to the epochs used for selecting the neurons for the next upcoming tasks, and {\tt Individual} refers to training a different model for each task. We can see that firefly descent learns the smallest network that achieves the best performance among all methods. Moreover, it is more 
computationally efficient than CPG when growing and picking the neurons for the new tasks. 
Figure~\ref{fig:cl}(b) shows the average accuracy over seen tasks on the fly. Again, firefly descent outperforms CPG by a significant margin.

\begin{table}[h]
\renewcommand\arraystretch{1.0}
    \centering
    \begin{tabular}{c|ccc}
    \hlineB{2.5}
        Method &  Param (M) & Extra Growing Epochs & Avg. Accuracy (20 tasks) \\
    \hline
          Individual  & 2565  & -   & 88.85
         \\
          CPG        & 289   & 420 & 90.75
         \\
          CPG w/o FC \footnotemark  & 28    & 420 & 90.58
         \\
          Firefly     & \textbf{26}    & \textbf{80}  & \textbf{91.03} \\
    \hlineB{2.5}
    \end{tabular}
    \vspace{5pt}
    \caption{20-way split continual image classification on CIFAR-100.} 
    \label{tab:cl20}
\end{table}
\footnotetext[4]{CPG without fully connected layers is to align the model structure and model size with {\tt Firefly}.}


\vspace{-15pt}
\section{Related Works}\label{sec:related}
In this section, we briefly review previous works that grow neural networks in a general purpose and then discuss existing works that apply network growing to tackle continual learning.

\paragraph{Growing for general purpose}
Previous works have investigated ways of knowledge transfer by expanding the network architecture. One of the approaches, called Net2Net \citep{wei2016network}, provides growing operations for widening and deepening the network with the same output. So whenever the network is applied to learn a new task, it will be initialized as a functional equivalent but larger network for more learning capacity. Network Morphism \citep{wei2016network} extends the Net2Net to a broader concept, which defines more operations that change a network's architecture but maintains its functional representation. Although the growing methods are similar to ours, in these works, they randomly or adopt simple heuristic to select which neurons to grow and in what direction. As a result, they failed to guarantee that the growing procedure can finally reach a better architecture every time. \citep{elsken2017simple} solve this problem by growing several neighboring networks and choose the best one after some training and evaluation on them. However, this requires comparing multiple candidate networks simultaneously.

On the other hand, recently, \citet{splitting2019} introduces the Splitting Steepest Descent, the first principled approach that determines which neurons to split and to where. By forming the splitting procedure into an optimization problem, the method finds the eigen direction of a local second-order approximation as the optimal splitting direction. However, the method is restricted to only splitting neurons. Generalizing it to special network structure requires case-by-case derivation and it is in general hard to directly apply it on other ways of growing. Moreover, since the method evaluates the second-order information at each splitting step, it is both time and space inefficient.

\citet{hu2019efficient} proposes an efficient path selection method by jointly training all the possible path, select the best subset and add to the network. However, it only discusses the situation in the cell-based network. Our adding strategy treats the general single neuron as the basic unit, thus, firefly can greatly extend the growing scheme into various of scenarios. 

\paragraph{Growing for continual learning} continual learning is a natural downstream application of growing neural networks. ProgressiveNet~\citep{rusu2016progressive} was one of the earliest to expand the neural network for learning new tasks while fixing the weights learned from previous tasks to avoid forgetting. LwF~\citep{li2017learning} divides the network into the shared and the task-specific parts, where the latter keeps branching for new tasks. Dynamic-expansion Net~\citep{yoon2017lifelong} further applies sparse regularization to make each expansion compact. Along this direction, \cite{hung2019increasingly,hung2019compacting} adopt pruning methods to better ensure the compactness of the grown model. All of these works use heuristics to expand the networks. By contrast, {\tt Firefly} is developed as a more principled growing approach. We believe future works can build theoretical analysis on top of the {\tt Firefly} framework.

\section{Conclusion}\label{sec:conclusion}
In this work, we present a simple but highly flexible framework for progressively growing neural networks in a principled steepest descent fashion. 
Our framework allows us to incorporate various mechanisms for growing networks (both in width and depth).
Furthermore, we demonstrate the effectiveness of our method on both growing networks on both single tasks and continual learning problems, in which our method consistently achieves the best results. Future work can investigate various other growing methods for specific applications under the general framework.  
\section*{Acknowledge}
The work is conducted in the statistical learning and AI group in computer science at UT Austin, which is supported in part by CAREER-1846421, SenSE-2037267, EAGER-2041327, and NSF AI Institute for Foundations of Machine Learning (IFML). In addition, this work's author Peter Stone is supported by CPS-1739964, IIS-1724157, ONR (N00014-18-2243), FLI (RFP2-000), ARO (W911NF-19-2-0333), DARPA, Lockheed Martin, GM, and Bosch. Peter Stone serves as the Executive Director of Sony AI America and receives financial compensation for this work. The terms of this arrangement have been reviewed and approved by the University of Texas at Austin in accordance with its policy on objectivity in research.

\bibliography{reference}
\bibliographystyle{nips}
\newpage\clearpage
\appendix
\section{Detailed Algorithm for Continual Learning}
Algorithm~\ref{alg:cl} summarizes the pipeline of applying firefly descent on growing neural architectures for continual learning problems.
\label{app:firefly}
\begin{algorithm*}[h] 
\caption{Firefly Steepest Descent for Continual  Learning} 
\begin{algorithmic} 
    \STATE \textbf{Input} : A stream of datasets $\{\mathcal{D}_1, \mathcal{D}_2, \dots, \mathcal{D}_T\}$;
    \FOR{task $t=1:T$}
        \IF{$t = 1$}
            \STATE Train $f_1$ on $\mathcal{D}_1$ for several epochs until convergence.
            \STATE Set mask $m_1$ to all $1$ vector over $f_1$.
        \ELSE
            \STATE Denote $f_t \leftarrow f_{1:{t-1}}$ and lock its weights.
            \STATE Train a binary mask $m_t$ over $f_t$ on $\mathcal{D}_t$ for several epochs until convergence.
        \ENDIF
        \STATE $f_t = f_t[m_t]$ \,\, \text{\small// $f_t$ is re-initialized as the selected old neurons from $f_{1:t-1}$ with their weights fixed.}
        \WHILE{$f_t$ can not solve task $t$ sufficiently well}
            \IF{$t = 1$}
                \STATE Grow $f_t$ by \textbf{splitting} existing neurons and growing new neurons.
            \ELSE
                \STATE Grow $f_t$ by \textbf{unlocking} existing neurons and growing new neurons.
            \ENDIF
            \STATE Train $f_t$ on $\mathcal{D}_t$
        \ENDWHILE
        \STATE Update $m_t$ as the binary mask over $f_t$.
        \STATE Record the network mask $m_t$, $f_{1:t} = f_{1:{1-t}}\cup f_t$.
    \ENDFOR
\end{algorithmic}
\label{alg:cl}
\end{algorithm*}

\section{Experiment Detail}
\subsection{Toy RBF Network}
We construct a following one-dimensional two-layer radial-basis function (RBF) neural network with one-dimensional inputs, 
\begin{equation}
\label{equ:rbf}
f(x) = \sum_{i=1}^{m} w_i \sigma(\theta_{i1} x + \theta_{i2}), \quad \text{where } ~~\sigma(t) = \exp (-\frac{t^2}{2}), ~~~~~x \in \RR,
\end{equation}
where $w_i \in \mathbb R$ and $\theta_i = [\theta_{1i},\theta_{2i}]$ are the input and output weights of the $i$-th neuron, respectively. We generate our true function by drawing $m = 15$ neurons with $w_i$ and $\theta_i$ i.i.d. from  $\normal(0,3)$.
For dataset $\{x^{(\ell)}, y^{(\ell)}\}_{\ell=1}^{1000}$, we generate them
with $x^{(\ell)}$ drawing from $\mathrm{Uniform}([-5,5])$ and let $y^{(\ell)} = f(x^{(\ell)})$.
We apply various growing methods to grow the network from one single neuron all the way up to 12 neurons.

For the new initialized neurons introduce during the growing in {\tt RandSearch} and {\tt Firefly}, we draw the neruons from $N(0,0.1)$.
For {\tt RandSearch}, we finetune all the randomly grow networks for 100 iterations.
For {\tt Firefly}, we also train the expanded network for 100 iterations before calculating the score and picking the neurons.
Further, We update 10,000 iterations between two consecutive growing.

\label{app:toy}
\subsection{Growing Wider and Deeper Networks}
\label{app:grow}
\paragraph{Setting for Growing Wider Networks}
For all the experiment including Net2Net, splitting steepest descent, NASH and our firefly descent, we grow 30\% more neurons each time. Between two consecutive grows, we finetune the network for 160 epochs.

For splitting steepest descent, we follow exactly the same setting as in \cite{splitting2019}.

For NASH, we only apply ``Network morphism Type II'' operation described in  \cite{elsken2017simple}, which is equivalent to growing the network width by randomly splitting the existing neurons.. During the search phase, we follow the original paper's setting, sample 8 neighbour networks, train each of them for 17 epochs and choose the best one as the grow result.

For firefly descent,  we grow a network by both splitting existing neurons and adding brand new neurons for widening the  network; When growing, we split all the existing neurons and add $m'=50$ brand new neurons draw from $N(0,0.1)$. We will also train the expanded network for 1 epoch before calculating the score and picking the neurons.

\paragraph{Growing Wider MobileNet V1}
We also compare firefly with other growing method on MobileNet V1 using CIFAR-100 dataset. Same as \cite{wu2020steepest}, we start from a thinner MobilNet V1 with 32 channels in each layer. We grow 35\% more neurons each time, the other settings are same as the previous growing wider networks' setting.

\begin{figure}[ht]
    \centering
    \includegraphics[width=1.0\textwidth]{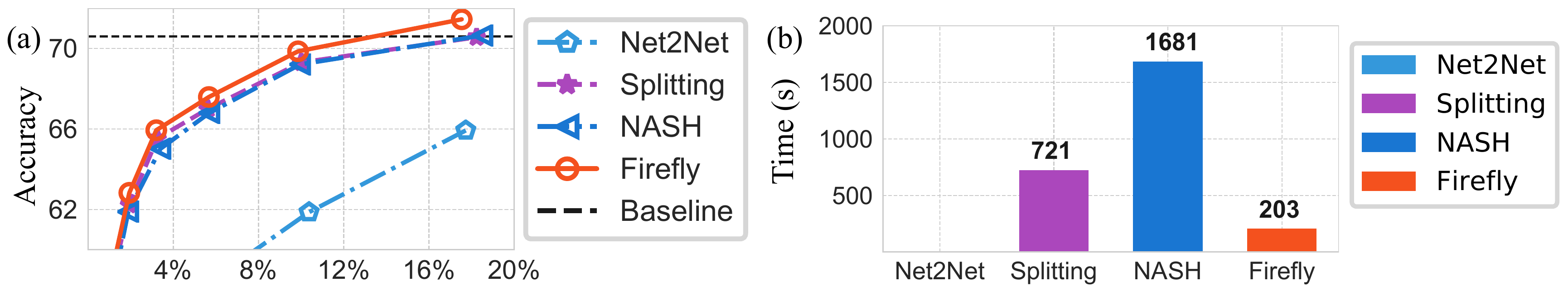}
    \caption{Results and time consumption of growing increasingly wider networks on CIFAR-100 using MobileNet V1 backbone}
    \label{fig:mbv1}
\end{figure}

Figure \ref{fig:mbv1} again shows that firefly splitting can out perform various of growing baseline on the same backbone network. Meanwhile, its time cost is much smaller than splitting and NASH algorithm.
\paragraph{Growing Deeper Networks}
We test firefly descent for growing network depth. We build a network with 4 blocks. Each block contains numbers of convolution layers with kernel size 3. The first convolution layer in each block is stride two. For a simple and clear explanation, we mark the number of layers in these 4 blocks as 12-12-12-12, for example, which means each block contains 12 layers.
Begin from 1-1-1-1, we grow the network using firefly descent on MNIST, FashionMNIST, SVHN, and compare it with AutoGrow \cite{wen2019autogrow} and NASH \cite{elsken2017simple}.

For our method, we start from a 1-1-1-1 network with 16 channels in each layer. We also insert 11 identity layers in each block, which roughly match the final number of layers in AutoGrow. We apply our growing layer strategy described in Section \ref{sec:grow_layers} for growing new layers and apply both splitting existing neurons and adding brand new neurons for widening the existing layers. When growing new layers, we introduce $m' = 20$ new neurons in each Identity map layers, when increasing the width of the existing layers, we split all the existing neurons and add $m' = 20$ new neurons. After expanding the network, we train the network for 1 epoch before calculating the score. If the Identity layer remains 2 or more new neurons after selection, we add this Identity layers in the network and train with the existing network together. Otherwise, we will remove all the new neurons and keep this layer as an Identity map. For the existing neurons, we grow 25\% of the total width.

For NASH, we apply ``Network morphism Type I'' and ``Network morphism Type II'' together, which represent growing depth by randomly insert identity layer and growing width by randomly splitting the existing neurons. During the search phase, we follow the original paper's setting, sample 8 neighbor networks, train each of them for 17 epochs and choose the best one as the growing result. Each time when sampling the neighbour networks, we grow the total width of the existing layers by $25\%$ and then randomly insert one layer in each blocks.

For both our method and NASH, we grow 11 steps and finetune 40 epochs after each grow step. We also retrain the searched network for 200 epochs after the last grow to get the final performance on each dataset.

For AutoGrow, we use the result report in the original paper.
\begin{table}[ht!]
\renewcommand\arraystretch{1.0}
\centering
\begin{tabular}{c|c|ccc}

\hlineB{2.5}
  Dataset& Method &Structure& Param (M) & Accuracy\\
\hline

\multirow{3}*{MNIST} & AutoGrow \cite{wen2019autogrow} & 13-12-12-12 & 2.3 & 99.57 \\

& NASH \cite{elsken2017simple} & 12-12-12-12 & 2.0 & 99.50 \\
& Firefly & 12-12-12-12 & \textbf{1.9} & \textbf{99.59} \\

\hline
\multirow{3}*{FashionMNIST} & AutoGrow \cite{wen2019autogrow} & 13-13-13-13 & 2.3 & 94.47 \\
& NASH \cite{elsken2017simple}& 12-12-12-12 & 2.2 & 94.34 \\
& Firefly & 12-12-12-12 & \textbf{2.1} & \textbf{94.48} \\
\hline
\multirow{3}*{SVHN} & AutoGrow \cite{wen2019autogrow} & 12-12-12-11 & 2.2 & 97.08 \\
& NASH \cite{elsken2017simple}& 12-12-12-12 & 2.0 & 96.90 \\
& Firefly & 12-12-12-12 & \textbf{1.9} & 97.08 \\
\hlineB{2.5}

\end{tabular}
\label{tab:depth}
\caption{Result on growing Depth comparing with two baselines }
\end{table}

Table \ref{tab:depth} shows the result. We can see our method can grow a smaller network to achieve the AutoGrow's performance and outperform the network searched with NASH.

\subsection{Application on Neural Architecture Search}
\label{app:nas}

Following the setting in DARTS \citep{liu2018darts}, we separate half of the CIFAR-10 training set as the validation set for growing. We start with a stacked 5 cell network for searching, the second and the fourth cell are reduction cells, which means all the operations next
to the input of the cells are set to stride two. In each cell, we build the SepConv and DilConv operation blocks following DARTS \citep{liu2018darts}. To apply our firefly descent, we grow the last convolution layer in each block and add a linear transform layer with the same output channels to ensure all the operations on the same edge can sum up in the same size as the output. The number of channels of the operations in each cell is set to 4-8-8-16-16, which is $0.25\times$ of that in the original Darts. The last linear transform layer in each cell has channels 16-32-32-64-64. We grow the network by both splitting existing neurons and adding brand new neurons, and each time we sequentially select one cell to grow. We repeat growing the whole 5 cells twice, which means we apply our firefly descent for 10 times in total. Each time, we split all the existing neurons in the chosen cell and add 4, 8, 8, 16, 16 brand new neurons differently for the 5 cells. We then train the expanded network for 5 epochs and select $25\%$ neurons to grow. As a result, we search the network structure for 100 epochs in total. All other training hyperparameters are set to the same values as in DARTS \citep{liu2018darts}.

After searching, we select the operation with the largest width in each edge as the final operation.  Besides, if all operations on the same edge grow less than $20\%$ comparing to the initial width, we assign this edge as Identity map in the final structure. We only keep the type of operations in the cell as our final search result because we need to increase the channel width to match the model size with the baselines.

For the final evaluation, we sequentially stack a 20 cell network and mark those cells as 1-20. We apply the search result of the first, second, third, fourth, and the fifth cell in the 5 stacked search network to cell 1-6, cell 7, cell 8-13, cell 14, and cell 15-20 of the final evaluation network accordingly. We increase the initial channel to 40 to match the model size with other baselines. The other training settings are kept the same as in DARTS \citep{liu2018darts}. Our result is averaged over 5 runs from our final evaluation model.

\subsection{Continual Learning}
\label{app:cl}
For both 10-way split CIFAR-100 and 20-way split CIFAR-100, we repeat the experiment 3 times with 3 different task splits. We apply both the copy-exist-neuron and grow-new-neuron strategies to tackle the CL problem. During each growing iteration, we add $15$ brand new neurons for each layer as candidates for growing. After expanding the network, we finetune the network for 50 epochs on the new task. During the selection phase, for 20-way split CIFAR-100, we select out the top 256 neurons among all the copied neurons and new neurons. For 10-way split CIFAR-100, we select the top ${32, 128, 196, 256, 320, 384, 448, 512}$ neurons each time to test our performance under different model size. After selecting the neurons, we finetune the expanded network on the new task for 100 epochs.

\end{document}